\documentclass{article}
\usepackage{spconf,amsmath,graphicx,hyperref,booktabs}

\title{Domain Recentering and Confidence-Weighted Prior Calibration \\for Vision-Language Models}
\name{Youngeun Seol$^{\star}$ \qquad Jimin Shin$^{\star}$ \qquad Heeseo Yoon$^{\star}$\thanks{$^*$ Equal contribution, listed in alphabetical order} \qquad Uiwon Hwang$^{\dagger}$\thanks{$^\dagger$Corresponding author: Uiwon Hwang (uiwon.hwang@ewha.ac.kr)}}
\address{Department of Computer Science and Engineering, Ewha Womans University}
\begin{document}
%
\maketitle
\begin{abstract}

Vision-language models such as CLIP achieve strong zero-shot classification, yet under distribution shift, visual embeddings drift from fixed text embeddings. Training-free calibration avoids the per-sample optimization of prompt learning, but prior feature calibration gives each image the full bias of one hard cluster. We propose \textbf{D}omain \textbf{R}ecentering with \textbf{C}onfidence Calibration (DRC), a training-free method adapting CLIP from a set of unlabeled target images. DRC fits a Gaussian mixture once and subtracts from each embedding a posterior-weighted average of component means. It then removes residual class preference with a log-prior correction, estimating the prior from confidence-weighted predictions. Among compared methods, DRC achieves the highest average accuracy on cross-domain datasets, exceeding zero-shot CLIP by 4.13 and 5.07 points with ViT-B/16 and ResNet-50, with gains over CLIP also holding under ImageNet distribution shifts.
\end{abstract}
\begin{keywords}
Vision-language models, domain recentering, confidence-weighted calibration
\end{keywords}

\section{Introduction}

Vision-language models have emerged as powerful foundation models for visual recognition. CLIP \cite{radford2021learning} learns a multi-modal embedding space for images and text from natural language supervision at scale, enabling zero-shot classification by aligning an image's embedding with the text embeddings of class descriptions. However, when target images differ from the pretraining distribution, their visual embeddings can become misaligned with the fixed text embeddings, degrading classification accuracy.

Recent works have adapted CLIP under distribution shift along two broad directions. One line performs prompt learning with few-shot labeled data \cite{zhou2022learning, zhou2022conditional} or test-time entropy minimization \cite{shu2022test, feng2023diverse}, requiring labels or per-image backpropagation. Another line calibrates features without training, using unlabeled target data. UMFC \cite{liang2024umfc} corrects visual encoder bias by subtracting from each feature the mean of its hard-assigned cluster. However, a feature near a cluster boundary then receives one cluster's full mean, so a small change in the feature can flip its correction. Correcting the image features alone also leaves the text classifier unchanged, so any preference it holds for certain classes persists.

We propose Domain Recentering with Confidence Calibration (DRC), which models the structure of unlabeled CLIP visual features using a Gaussian mixture model. Rather than assigning each image to a single cluster, DRC uses mixture-component posterior probabilities to construct a soft, image-specific bias vector as a weighted combination of component means, so the correction changes continuously as the feature moves between components. This vector is subtracted from the image feature to suppress domain-related variation while preserving class-discriminative content. DRC further corrects residual class preferences through confidence-weighted prior calibration, without any backpropagation or parameter updates. Among the compared methods, DRC achieves the highest average accuracy on ten cross-domain datasets with both ResNet-50 and ViT-B/16, and it also improves the average accuracy of zero-shot CLIP over four ImageNet distribution shifts.
\section{Related Work}
Prompt learning adapts CLIP \cite{radford2021learning} to downstream data by tuning its textual prompts. CoOp \cite{zhou2022learning} and CoCoOp \cite{zhou2022conditional} learn these prompts from labeled few-shot data. Test-time adaptation instead uses unlabeled test samples to improve robustness under distribution shifts. TPT \cite{shu2022test} optimizes prompts for each test sample, and DiffTPT \cite{feng2023diverse} extends TPT with diffusion-based augmentation. Both repeat backpropagation for every test sample, which makes inference costly and motivates training-free adaptation that leaves the pretrained model unchanged.

Training-free methods instead correct the systematic biases that zero-shot vision-language models exhibit in representations and predictions \cite{liang2024umfc, parashar2024neglected}. UMFC \cite{liang2024umfc} subtracts each hard-assigned cluster's mean, giving all images within a cluster the same correction. Label-free logit adjustment \cite{zhu2024enhancing} offsets label bias in predictions. Our method instead assigns each image its own bias by weighting Gaussian mixture component means with the image's soft assignment. It then estimates the class prior from the recentered predictions with confidence weighting, so that the prior targets the bias left after feature correction.
\section{Method}

\subsection{Background}

\textbf{CLIP.}
CLIP \cite{radford2021learning} maps images and text into a shared embedding space via a visual encoder $E_v$ and a text encoder $E_t$. For zero-shot classification, each class name $y_c$ is inserted into prompt templates and encoded by $E_t$, and the averaged embedding serves as the classifier weight $\mathbf{t}_c$. Given an image $x_i$, its visual feature $\mathbf{v}_i = E_v(x_i)$ is compared to each $\mathbf{t}_c$ via scaled cosine similarity
\begin{equation}
\ell_{ic} =
\exp(\tau)
\left(\frac{\mathbf{v}_i}{\lVert\mathbf{v}_i\rVert_2}\right)^{\top}
\frac{\mathbf{t}_c}{\lVert\mathbf{t}_c\rVert_2},
\end{equation}
where $\exp(\tau)$ is the learned logit scale of CLIP, clipped at $100$. CLIP predicts the class with the highest logit $\ell_{ic}$, and under domain shift, $\mathbf{v}_i$ can become misaligned with the fixed $\mathbf{t}_c$. \\

\begin{figure}
    \centering
    \includegraphics[width=1\linewidth]{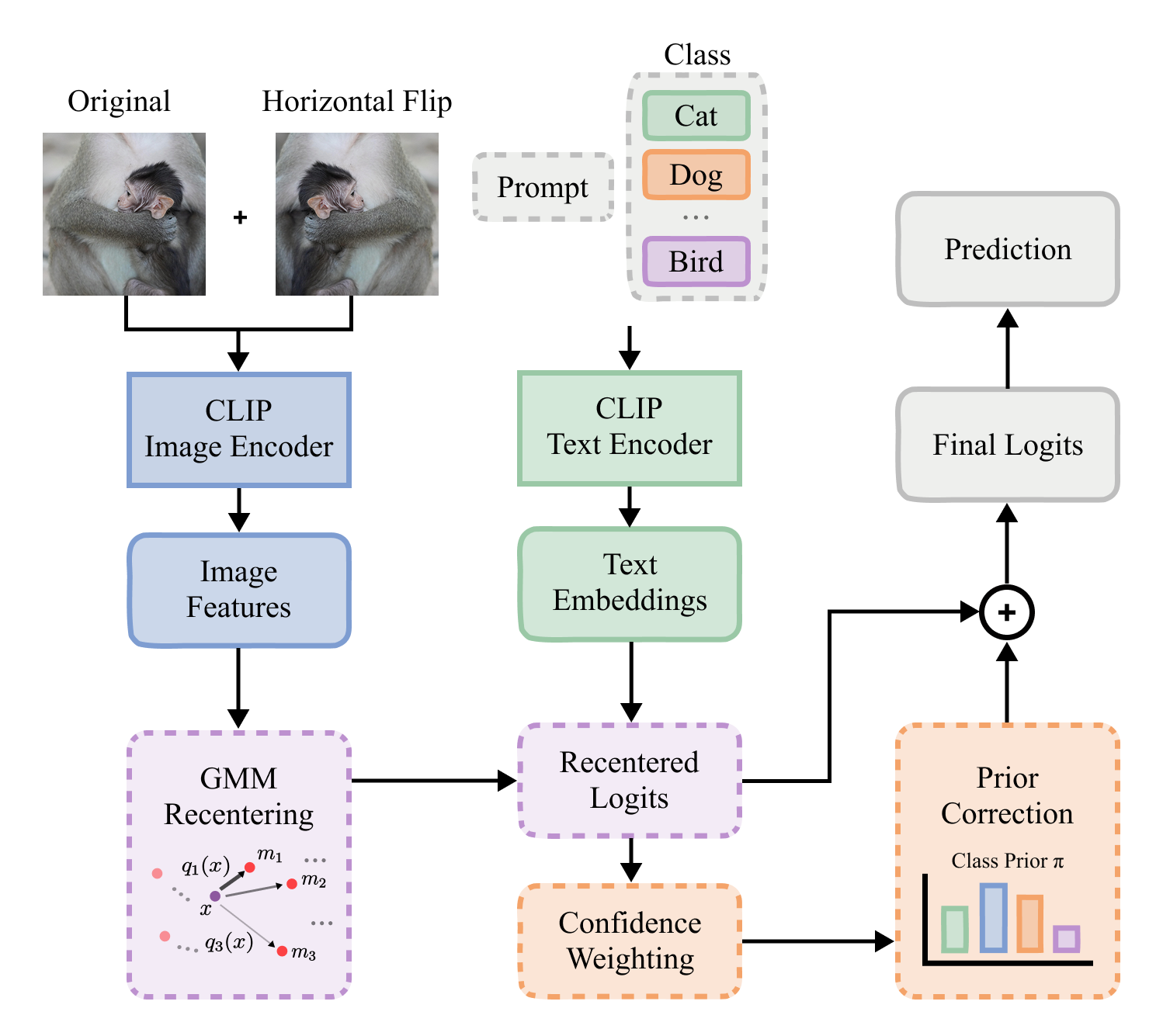}
    \caption{Overview of DRC. DRC recenters CLIP image features with a GMM-based soft bias correction and calibrates the resulting logits with a confidence-weighted class prior.}
    \label{fig:pipeline}
\end{figure}

\noindent\textbf{Problem Setting.}
We adapt a frozen CLIP model to a target domain whose images may differ from the pretraining distribution. An unlabeled adaptation set $\mathcal{D}_u = \{x_i\}_{i=1}^{N}$ of target-domain images is given. DRC estimates its statistics once from $\mathcal{D}_u$ without any target labels and then classifies each test image with these fixed statistics. The visual encoder $E_v$, the text encoder $E_t$, and the classifier weights $\mathbf{t}_c$ remain unchanged throughout. Figure \ref{fig:pipeline} illustrates the DRC pipeline, which recenters each image feature with a Gaussian mixture fitted on $\mathcal{D}_u$ and then calibrates the logits with a confidence-weighted class prior.

\subsection{Domain Recentering}
DRC models the features of $\mathcal{D}_u$ with a Gaussian mixture and estimates the bias of each image from all mixture components, weighting each component by how strongly the image belongs to it.

Each image, whether in $\mathcal{D}_u$ or at test time, is represented by averaging its $\ell_2$-normalized feature with that of its horizontal flip, followed by re-normalization
\begin{equation}
\mathbf{f}_i = \operatorname{norm}\big(\operatorname{norm}(E_v(x_i)) + \operatorname{norm}(E_v(\operatorname{Flip}(x_i)))\big).
\end{equation}

With the mean $\boldsymbol{\mu} = \frac{1}{N}\sum_{i=1}^{N}\mathbf{f}_i$ over $\mathcal{D}_u$, the centered features are projected via PCA onto $\min(16,D,N)$ dimensions, where $D$ is the CLIP embedding dimension, so that the mixture is fitted on the dominant directions of variation
\begin{equation}
\mathbf{z}_i = \mathbf{P}(\mathbf{f}_i - \boldsymbol{\mu}).
\end{equation}

A diagonal-covariance GMM with $K$ components,
\begin{equation}
p(\mathbf{z}) = \sum_{k=1}^{K} \pi_k \mathcal{N}(\mathbf{z}; \boldsymbol{\nu}_k, \mathbf{\Sigma}_k),
\end{equation}
is fitted to $\{\mathbf{z}_i\}$, where $\pi_k$, $\boldsymbol{\nu}_k$, and $\mathbf{\Sigma}_k$ denote the mixture weight, mean, and diagonal covariance of component $k$. Unlike hard clustering such as K-means, the GMM assigns each feature softly to the $K$ components through its posterior $q_{i,k} = p(k \mid \mathbf{z}_i)$.

Each component mean is computed in the original embedding space from the features whose most probable component is $k$, $\mathcal{I}_k = \{i \mid k = \arg\max_j q_{i,j}\}$, so that it summarizes the features belonging mainly to that component
\begin{equation}
\mathbf{m}_k = \frac{1}{|\mathcal{I}_k|}\sum_{i \in \mathcal{I}_k}\mathbf{f}_i.
\end{equation}
For an input feature $\mathbf{f}$, DRC estimates its bias by weighting these means with the posterior
\begin{equation}
\mathbf{b}(\mathbf{f}) = \sum_{k=1}^{K} q_k(\mathbf{f})\,\mathbf{m}_k, \qquad q_k(\mathbf{f}) = p\big(k \mid \mathbf{P}(\mathbf{f}-\boldsymbol{\mu})\big).
\end{equation}
Since $q_k(\mathbf{f})$ varies continuously with $\mathbf{f}$, the bias changes gradually as a feature moves between components, whereas a hard assignment would switch it abruptly at component boundaries.

The recentered feature is obtained by subtracting this bias and re-normalizing
\begin{equation}
\hat{\mathbf{f}} = \frac{\mathbf{f} - \beta\,\mathbf{b}(\mathbf{f})}{\left\|\mathbf{f} - \beta\,\mathbf{b}(\mathbf{f})\right\|_2},
\end{equation}
where $\beta$ controls the strength of recentering.

\subsection{Confidence-Weighted Prior Calibration}
Recentering acts only on the image features and leaves the classifier weights $\mathbf{t}_c$ unchanged, so the class preference of the zero-shot classifier \cite{zhu2024enhancing, parashar2024neglected} can persist in the recentered predictions. DRC estimates this preference from predictions on $\mathcal{D}_u$ and removes it with a log-prior adjustment, weighting each prediction by its confidence.

For each $x_i \in \mathcal{D}_u$, let $\hat{\ell}_{ic}$ be the logit from the recentered feature. The class probability is then
\begin{equation}
p_i(c) = \frac{\exp(\hat{\ell}_{ic})}{\sum_{c'} \exp(\hat{\ell}_{ic'})}.
\end{equation}

Since an uncertain prediction provides weak evidence of class preference, each prediction is weighted by one minus its normalized entropy
\begin{equation}
H_i = -\sum_c p_i(c)\log p_i(c),
\qquad
w_i = 1 - \frac{H_i}{\log C},
\end{equation}
where $C$ is the number of classes, so that near-uniform predictions receive weights close to zero. The confidence-weighted class prior is 
\begin{equation}
\pi_c^{\mathrm{Conf}} = \frac{\sum_{i=1}^{N} w_i\, p_i(c)}{\sum_{i=1}^{N} w_i}.
\end{equation}
The computed distribution summarizes the class preference reflected in reliable predictions on the adaptation set. A large $\pi_c^{\mathrm{Conf}}$ indicates that class $c$ receives a relatively large amount of prediction mass, while a small value denotes that the class is less frequently preferred by the model. Confidence weighting reduces the influence of uncertain predictions that could otherwise distort this estimate.

The log-prior correction and its zero-centered form are
\begin{equation}
r_c = -\log(\pi_c^{\mathrm{Conf}} + \epsilon), \qquad \tilde{r}_c = r_c - \frac{1}{C}\sum_{c'=1}^{C} r_{c'},
\end{equation}
where $\epsilon$ is a small constant for numerical stability. 
A test image with recentered feature $\hat{\mathbf{f}}$ is classified with the calibrated logits
\begin{equation}
\ell'_{c} = \hat{\ell}_{c} + \tilde{r}_c, \qquad \hat{\ell}_{c} = \exp(\tau)\,\hat{\mathbf{f}}^{\top}\frac{\mathbf{t}_c}{\lVert\mathbf{t}_c\rVert_2}.
\end{equation}
This adjustment lowers the logits of classes that receive excessive prediction mass on $\mathcal{D}_u$ and raises those of less preferred classes. 
\section{Experiments}

\subsection{Experimental Setup}

\textbf{Datasets.}
We evaluate DRC on the cross-domain benchmark and the out-of-distribution (OOD) benchmark. For cross-domain evaluation, we use the same ten datasets as TPT \cite{shu2022test}, which are Aircraft, Caltech101, Cars, DTD, EuroSAT, Flower102, Food101, Pets, SUN397, and UCF101. The OOD benchmark measures robustness to distribution shifts with four variants of ImageNet \cite{deng2009imagenet}, ImageNet-A \cite{hendrycks2021natural}, ImageNet-V2 \cite{recht2019imagenet}, ImageNet-R \cite{hendrycks2021many} and ImageNet-Sketch \cite{wang2019learning}. \\

\begin{table*}[!t]
\centering
\footnotesize
\setlength{\tabcolsep}{4pt}
\begin{tabular}{lccccccccccc}
\toprule
Method & Aircraft & Caltech101 & Cars & DTD & EuroSAT & Flower102 & Food101 & Pets & SUN397 & UCF101 & Average \\
\midrule
CLIP-ResNet-50 & 16.11 & 87.26 & 55.89 & 40.37 & 25.79 & 62.77 & 74.82 & 82.97 & 60.85 & 59.48 & 56.63 \\
\midrule
CoOp   & 15.12 & 86.53 & 55.32 & 37.29 & 26.20 & 61.55 & 75.59 & 87.00 & 58.15 & 59.05 & 56.18 \\
CoCoOp & 14.61 & \underline{87.38} & 56.22 & 38.53 & 28.73 & \underline{65.57} & 76.20 & \textbf{88.39} & 59.61 & 57.10 & 57.23 \\
\midrule
TPT     & 17.58 & 87.02 & 58.46 & 40.84 & 28.33 & 62.69 & 74.88 & 84.49 & 61.46 & 60.82 & 57.66 \\
DiffTPT & 17.60 & 86.89 & \textbf{60.71} & 40.72 & \underline{41.04} & 63.53 & \textbf{79.21} & 83.40 & \textbf{62.72} & 62.67 & \underline{59.85} \\
UMFC    & \underline{17.76} & 85.48 & 56.25 & \underline{42.85} & 37.63 & \underline{65.57} & 77.47 & 86.10 & 60.49 & \underline{63.39} & 59.30 \\
\textbf{DRC (Ours)} & \textbf{18.81} & \textbf{87.99} & \underline{59.66} & \textbf{44.33} & \textbf{44.84} & \textbf{66.30} & \underline{78.83} & \underline{88.01} & \underline{62.46} & \textbf{65.72} & \textbf{61.70} \\
\midrule
\midrule
CLIP-ViT-B/16 & 23.22 & 93.55 & 66.11 & 45.04 & 50.42 & 66.99 & 82.86 & 86.92 & 65.63 & 65.16 & 64.59 \\
\midrule
CoOp   & 18.47 & 93.70 & 64.51 & 41.92 & 46.39 & 68.71 & 85.30 & 89.14 & 64.15 & 66.55 & 63.88 \\
CoCoOp & 22.29 & \underline{93.79} & 64.90 & 45.45 & 39.23 & \underline{70.85} & 83.97 & \underline{90.46} & \underline{66.89} & 68.44 & 64.63 \\
\midrule
TPT     & 24.78 & \textbf{94.16} & 66.87 & \textbf{47.75} & 42.44 & 68.98 & 84.67 & 87.79 & 65.50 & 68.04 & 65.10 \\
DiffTPT & \underline{25.60} & 92.49 & \underline{67.01} & 47.00 & 43.13 & 70.10 & \textbf{87.23} & 88.22 & 65.74 & 68.22 & 65.47 \\
UMFC    & 25.47 & 91.44 & 66.35 & 45.57 & \underline{50.98} & 70.60 & 85.99 & 88.91 & 65.73 & \underline{68.91} & \underline{66.00} \\
\textbf{DRC (Ours)} & \textbf{27.39} & 92.17 & \textbf{68.82} & \underline{47.46} & \textbf{61.73} & \textbf{71.78} & \underline{86.80} & \textbf{91.11} & \textbf{67.52} & \textbf{72.43} & \textbf{68.72} \\
\bottomrule
\end{tabular}
\caption{Results on the Cross-Domain Benchmark with CLIP-ResNet-50 and CLIP-ViT-B/16. CoOp and CoCoOp serve as supervised transfer baselines trained on ImageNet with 16 labeled samples per class. CLIP, CoOp, CoCoOp, and TPT results are taken from the original TPT paper \cite{shu2022test}, and DiffTPT results are taken from the original DiffTPT paper \cite{feng2023diverse}.}
\label{tab:crossdomain}
\end{table*}

\begin{table*}[!t]
\centering
\footnotesize
\setlength{\tabcolsep}{6pt}
\begin{tabular}{lccccccc}
\toprule
Method & ImageNet & -A & -V2 & -R & -S & Average & OOD Average \\
\midrule
CLIP-ResNet-50 & 59.81 & 23.24 & 52.91 & 60.72 & 35.48 & 46.43 & 43.09 \\
\midrule
CoOp   & \textbf{63.33} & 23.06 & 55.40 & 56.60 & 34.67 & 46.61 & 42.43 \\
CoCoOp & \underline{62.81} & 23.32 & \underline{55.72} & 57.74 & 34.48 & 46.81 & 42.82 \\
\midrule
TPT     & 60.74 & \underline{26.67} & 54.70 & 59.11 & 35.09 & 47.26 & 43.89 \\
DiffTPT & 60.80 & \textbf{31.06} & \textbf{55.80} & 58.80 & \underline{37.10} & \textbf{48.71} & \textbf{45.69} \\
UMFC    & 59.70 & 23.47 & 52.85 & \underline{60.94} & 35.82 & 46.56 & 43.27 \\
\textbf{DRC (Ours)} & 62.12 & 23.77 & 55.44 & \textbf{61.39} & \textbf{39.94} & \underline{48.53} & \underline{45.14} \\
\midrule
\midrule
CLIP-ViT-B/16 & 68.34 & 49.89 & 61.88 & 77.65 & 48.24 & 61.20 & 59.42 \\
\midrule
CoOp   & \textbf{71.51} & 49.71 & \underline{64.20} & 75.21 & 47.99 & 61.72 & 59.28 \\
CoCoOp & \underline{71.02} & 50.63 & 64.07 & 76.18 & 48.75 & 62.13 & 59.91 \\
\midrule
TPT     & 68.98 & \underline{54.77} & 63.45 & 77.06 & 47.94 & \underline{62.44} & \underline{60.81} \\
DiffTPT & 70.30 & \textbf{55.68} & \textbf{65.10} & 75.00 & 46.80 & 62.28 & 60.52 \\
UMFC    & 68.21 & 50.33 & 61.99 & \underline{77.98} & \underline{48.77} & 61.46 & 59.77 \\
\textbf{DRC (Ours)} & 70.17 & 49.71 & 63.83 & \textbf{78.14} & \textbf{52.09} & \textbf{62.79} & \textbf{60.94} \\
\bottomrule
\end{tabular}
\caption{Results on the OOD Benchmark with CLIP-ResNet-50 and CLIP-ViT-B/16. CoOp and CoCoOp serve as supervised transfer baselines trained on ImageNet with 16 labeled samples per class. CLIP, CoOp, CoCoOp, and TPT results are taken from the original TPT paper \cite{shu2022test}, and DiffTPT results are taken from the original DiffTPT paper \cite{feng2023diverse}.}
\label{tab:ood}
\end{table*}

\noindent\textbf{Implementation details.}
Text embeddings obtained from dataset-specific prompt templates are averaged into a single vector for each class, and image features are extracted from each image and its horizontal flip. DRC uses PCA and a diagonal-covariance GMM ($n_{\mathrm{init}}=5$, seed 42). The adaptation set $\mathcal{D}_u$ is the validation split of each cross-domain dataset and the unlabeled evaluation images for ImageNet and its variants. $K$ and $\beta$ are selected by accuracy on the validation split of each cross-domain dataset, and on the ImageNet validation set for the OOD benchmark, with the ImageNet values reused for all four variants.

\subsection{Comparison with State-of-the-Art}

\textbf{Cross-Domain Benchmark.}
Table \ref{tab:crossdomain} evaluates DRC across ten datasets with disjoint class spaces. DRC achieves the highest average accuracy with both CLIP-ViT-B/16 and CLIP-ResNet-50, reaching 68.72\% and 61.70\%, respectively. These results improve over zero-shot CLIP by 4.13 and 5.07 points, and DRC ranks first on 7 and 6 of the 10 datasets, respectively. On EuroSAT, DRC improves over the best competing method by 10.75 and 3.80 points, and on UCF101, the gains are 3.52 and 2.33 points, respectively. \\

\noindent\textbf{OOD Benchmark.}
Table \ref{tab:ood} further evaluates DRC on the OOD benchmark. With CLIP-ViT-B/16, DRC achieves the highest Average and OOD Average of 62.79\% and 60.94\%. With CLIP-ResNet-50, DRC achieves 48.53\% and 45.14\%, improving over zero-shot CLIP by 2.10 and 2.05 points, respectively. The gains are particularly pronounced under strong appearance shifts. DRC achieves the highest accuracy on ImageNet-R/S for both backbones (61.39\%/39.94\% for CLIP-ResNet-50, 78.14\%/52.09\% for CLIP-ViT-B/16).

\begin{table}[t]
\centering
\footnotesize
\setlength{\tabcolsep}{10pt}
\begin{tabular}{lcc}
\toprule
& \multicolumn{2}{c}{Prior calibration} \\
\cmidrule(lr){2-3}
Recentering & without & with \\
\midrule
None & 64.59 & 68.35 \\
Hard & 66.89 & 68.57 \\
Soft & 67.01 & \textbf{68.72} \\
\bottomrule
\end{tabular}
\caption{Ablation on the cross-domain benchmark with CLIP-ViT-B/16, reporting the average accuracy over the ten datasets. The bottom-right entry is DRC.}
\label{tab:ablation}
\end{table}

\subsection{Ablation Study}

We perform ablation studies on the cross-domain benchmark with CLIP-ViT-B/16 to examine the effectiveness of domain recentering and confidence-weighted prior calibration.

As shown in Table \ref{tab:ablation}, both components improve over the CLIP baseline. Soft recentering outperforms hard recentering both without and with prior calibration. Confidence-weighted prior calibration provides the largest gain, increasing the average accuracy to 68.35\%. Combining both components yields the best result, with DRC achieving an average accuracy of 68.72\%.

\section{Conclusion}
In this paper, we proposed DRC, a training-free method that improves the robustness of CLIP under distribution shift without any backpropagation or parameter updates. DRC fits a Gaussian mixture to unlabeled target features and subtracts from each feature a soft, image-specific bias, so that the correction varies continuously across mixture components. It then corrects class preferences remaining in the recentered predictions with a confidence-weighted class prior. DRC achieves the highest average accuracy among the compared methods on the cross-domain benchmark and improves the OOD Average of zero-shot CLIP with both backbones. 

\newpage
\section{Acknowledgements}
This work was supported by the National Research Foundation of Korea (NRF) grant funded by the Korea government (MSIT) (RS-2025-00561169), Global – Learning \& Academic research institution for Master’s · PhD students, and Postdocs (G-LAMP) Program of the National Research Foundation of Korea (NRF) grant funded by the Ministry of Education (No. RS-2025-25442252), and Institute of Information \& communications Technology Planning \& Evaluation (IITP) under the Leading Generative AI Human Resources Development (IITP-2027-RS-2026-25546026) grant funded by the Korea government (MSIT).

\bibliographystyle{IEEEbib}
\bibliography{strings,refs}

\end{document}